\documentclass[runningheads]{llncs}
\usepackage{url}
\usepackage{pifont}
\usepackage[T1]{fontenc}
\usepackage{microtype}
\usepackage{graphicx,verbatim}
\usepackage{amsmath,amssymb}
\usepackage[table]{xcolor}
\definecolor{crossgray}{gray}{0.55}
\usepackage{algorithm}
\usepackage{algpseudocode}
\usepackage{booktabs}
\usepackage{threeparttable}
\usepackage{capt-of}
\definecolor{commentgray}{gray}{0.5}

\newcommand{\ms}[2]{#1${\pm}$#2}
\newcommand{\msb}[2]{\textbf{#1}${\pm}$\textcolor{beststd}{\textbf{#2}}}
\newcommand{\msp}[2]{\textbf{#1}${\pm}$#2}

\begin{document}
\title{MAGEFormer: Learning Metric-Consistent Representations for Anisotropic CT Segmentation}
\titlerunning{MAGEFormer}
\author{Jiaying Li \and Paolo Remagnino}
\authorrunning{J. Li and P. Remagnino}
\institute{Department of Computer Science, Durham University, Durham, UK\\
\email{jiaying.li2@durham.ac.uk}, \email{paolo.remagnino@durham.ac.uk}}

\maketitle              \begin{abstract} Vision Transformers (ViTs) have shown strong performance in volumetric segmentation, but their effectiveness on clinical CT is limited by an isotropic Euclidean lattice assumption. This conflicts with anisotropic CT acquisition, leading to two key issues: (1) a metric mismatch between voxel indices and physical anatomy, and (2) accuracy degradation from isotropic resampling. To address this, we propose MAGEFormer, a geometry-calibrated framework that embeds physical metric constraints directly into representation learning. Our method introduces Metric-Adaptive Spatial Embedding (MASE) to calibrate positional frequencies using voxel spacing, Geometry-Constrained Attention (GCA) to suppress physically implausible feature correlations, and Geometric View Voting (GVV) to reduce discretization bias during inference. We evaluate MAGEFormer on two multi-organ abdominal CT benchmarks, BTCV and FLARE\,22, under a unified protocol against strong CNN and Transformer-based baselines. MAGEFormer achieves the strongest boundary accuracy among the compared methods, with 10.58 mm HD95 on BTCV and 3.40 mm HD95 on FLARE\,22, and shows consistent gains in Dice under the same protocol. These results show that geometry-aware internal calibration is more effective than relying on conventional isotropic preprocessing alone for anisotropic CT segmentation.


\end{abstract}

\section{Introduction}
\label{sec:intro}
The rapid adoption of Vision Transformers (ViTs)~\cite{NIPS2017_3f5ee243,dosovitskiy2021an} has advanced volumetric medical image segmentation by improving long-range dependency modeling over traditional CNNs~\cite{2015_UNet}, with architectures such as nnFormer~\cite{nnformer_2023} further integrating convolution and self-attention for 3D volumes. However, most ViT-based architectures implicitly assume an isotropic Euclidean lattice ($\Delta x \approx \Delta y \approx \Delta z$), a geometric inductive bias that is often misaligned with the acquisition physics of clinical Computed Tomography (CT).\\
\indent In routine abdominal CT, the through-plane spacing ($s_z$) is frequently much coarser than the in-plane spacing ($s_{xy}$) due to radiation dose and acquisition speed constraints~\cite{RPLHR_CT,nnu-net_2021}, making the data inherently anisotropic. Without explicit calibration, self-attention may treat a physically distant voxel along the sparse $z$-axis as equally local to a neighbor in the dense $xy$-plane. This metric mismatch can hinder scale-consistent representation learning, especially for small or low-contrast structures (e.g., adrenal glands, pancreas tail) whose boundary quality is highly sensitive to sampling geometry~\cite{GE2023104590}.\\
\indent Existing methods address this issue in two main ways. Implicit adaptation via learnable or input-conditioned positional encodings~\cite{chu2021conditional} and relative position biases~\cite{9710917} (e.g., Swin-UNETR~\cite{swinunetr22}) is effective in practice, yet these are parameterized on voxel indices rather than physical distances and do not explicitly enforce metric consistency. Explicit architectural adaptation, such as thickness-aware operator design~\cite{10.1007/978-3-030-59719-1_55}, adapts network behavior to slice thickness but does not embed metric constraints into the learned representations themselves. Our goal is complementary: we treat metric consistency as a representation-level design principle that benefits from being enforced across encoding, feature interaction, and inference, rather than addressed solely through isotropic preprocessing or localized architectural heuristics. Throughout, by isotropic preprocessing we mean not a separately tuned baseline but the conventional resampling/normalization setting in which models operate on voxel-index neighborhoods as if the lattice were geometrically uniform.\\
\indent We propose MAGEFormer, a geometry-calibrated framework that models anisotropy as an intrinsic physical property within the representation learning process. It integrates three components: Metric-Adaptive Spatial Embedding (MASE), which calibrates positional frequencies using voxel spacing so that positional encoding reflects physical distance; Geometry-Constrained Attention (GCA), which reformulates skip-feature interaction with a metric-aware geometric bias to suppress physically implausible correlations; and Geometric View Voting (GVV), which reduces discretization bias during inference via deterministic sub-voxel view aggregation. Under a unified evaluation protocol on the BTCV and FLARE\,22 benchmarks, MAGEFormer achieves the strongest boundary accuracy among compared CNN and Transformer-based baselines, with consistent gains in Dice. Moreover, its advantage increases with anisotropy severity, supporting our central hypothesis that internal geometry calibration is more effective than relying on isotropic preprocessing alone.

\section{Method}
\label{sec:method}

\begin{figure*}[t] 
  \centering
  \includegraphics[width=\linewidth]{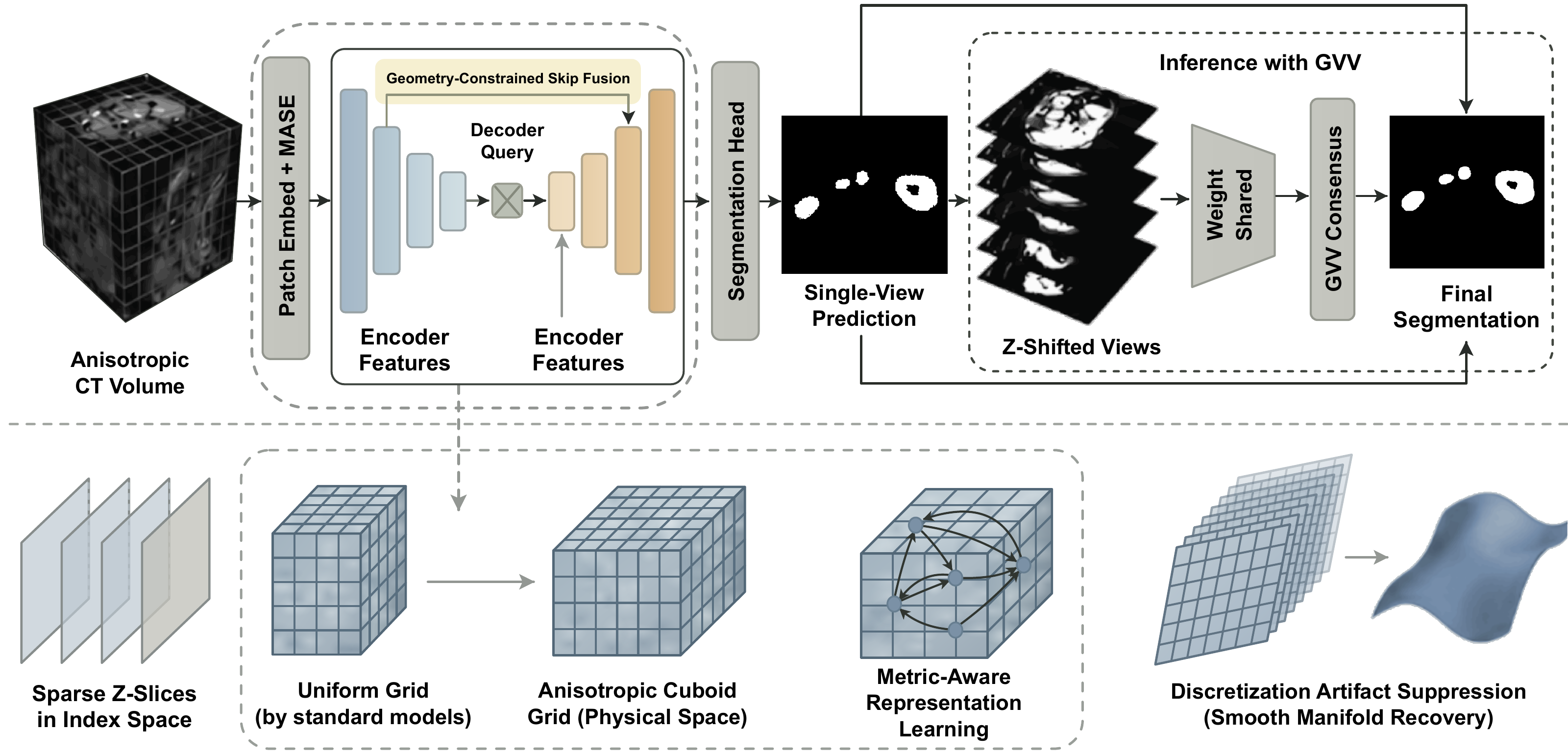}
  \caption{MAGEFormer: Architecture Overview and Geometric Motivation}
  \label{fig:architecture}
\end{figure*}

Metric consistency, the requirement that learned representations reflect physical distances rather than voxel indices, must be enforced at multiple pipeline stages. We operationalize this in MAGEFormer, a geometry-calibrated framework with three components addressing metric consistency at complementary levels: Metric-Adaptive Spatial Embedding (MASE) at positional encoding, Geometry-Constrained Attention (GCA) at cross-scale feature interaction, and Geometric View Voting (GVV) at inference-time discretization. An overview is shown in Fig.~\ref{fig:architecture}.

\subsection{Problem Formulation: Metric Mismatch in Anisotropic Grids}
\label{subsec:formulation}
Let a CT volume be sampled on a lattice $\Lambda \subset \mathbb{Z}^{3}$ with voxel index $\mathbf{p}=(p_x,p_y,p_z)$ and spacing
$\mathbf{s}^{(0)}=(s_x^{(0)},s_y^{(0)},s_z^{(0)})$, typically anisotropic ($s_z^{(0)} \gg s_x^{(0)}, s_y^{(0)}$), yet most models implicitly assume isotropic geometry. After reorientation to a canonical frame, the physical metric is $G^{(0)}=\mathrm{diag}((s_x^{(0)})^2,(s_y^{(0)})^2,(s_z^{(0)})^2)$ and the physically meaningful distance is
\begin{equation}
    d_{\mathcal{S}}(\mathbf{p}, \mathbf{q})
    = \sqrt{(\mathbf{p}-\mathbf{q})^\top G^{(0)}(\mathbf{p}-\mathbf{q})}.
    \label{eq:physical_distance}
\end{equation}
Equal index offsets thus do not imply equal physical displacements, especially along the sparse through-plane axis.

MAGEFormer adopts a multi-scale U-shaped Transformer backbone. The mismatch propagates through the hierarchy: at stage $\ell$, one token spans multiple voxels with effective spacing $\mathbf{s}^{(\ell)} = \mathbf{m}^{(\ell)} \odot \mathbf{s}^{(0)}$, where $\mathbf{m}^{(\ell)}$ is the cumulative stride. This scale-dependent metric forms the shared geometric substrate on which all three components operate.

\subsection{MASE: Metric-Adaptive Spatial Embedding}
\label{subsec:mase}

Standard positional encodings, including learnable and rotary variants~\cite{SU2024127063,heo2024rotary}, parameterize phase on index offsets and thus cannot distinguish voxels close in index space but distant physically. MASE reformulates positional phase in a spacing-normalized coordinate system so that phase progression tracks physical displacement. Following the principle that Fourier feature mappings resolve high-frequency spatial variations~\cite{NEURIPS2020_55053683}, we realize this via an axis-wise rotary form.

At stage $\ell$, with in-plane reference scale $s_{\mathrm{ref}}^{(\ell)}=\sqrt{s_x^{(\ell)}s_y^{(\ell)}}$, we define
\begin{align}
    \rho_a^{(\ell)} &= \frac{s_a^{(\ell)}}{s_{\mathrm{ref}}^{(\ell)}}\;\;(a\in\{x,y,z\}),
    \qquad
    \gamma^{(\ell)} = \sigma(\hat{\gamma}^{(\ell)})\;\;(\gamma^{(\ell)}\in(0,1)), \label{eq:rho}\\
    \tilde{\omega}_{k}^{(\ell,a)} &= \omega_k \rho_a^{(\ell)},
    \qquad
    \tilde{\omega}_{z,k}^{(\ell)} = \omega_k \left[1 + \gamma^{(\ell)}\big(\rho_z^{(\ell)} - 1\big)\right]. \label{eq:z_freq_refined}
\end{align}
where $\omega_k$ is the base angular frequency and $\sigma(\cdot)$ the sigmoid. We initialize $\hat{\gamma}^{(\ell)}$ so that $\gamma^{(\ell)} \approx 1$, recovering full metric correction. The learnable refinement is restricted to the $z$-axis---the dominant source of mismatch in abdominal CT---letting the network interpolate between strict physical calibration and a data-driven optimum.

For each head, channels are partitioned into axis-specific subspaces ($x/y/z$), independently transformed, and concatenated. For axis $a$ and token coordinate $p_a$:
\begin{equation}
\left(u^{(a)}_{2k-1} + i u^{(a)}_{2k}\right)
\mapsto
\left(u^{(a)}_{2k-1} + i u^{(a)}_{2k}\right)
\exp\!\left(i\, p_a\, \tilde{\omega}^{(\ell,a)}_k\right),
\label{eq:mase_rotation_axis}
\end{equation}
where $\tilde{\omega}^{(\ell,z)}_k$ uses Eq.~\eqref{eq:z_freq_refined} and $\tilde{\omega}^{(\ell,x)}_k,\tilde{\omega}^{(\ell,y)}_k$ use Eq.~\eqref{eq:rho}--\eqref{eq:z_freq_refined}. MASE is applied to query and key streams in all self-attention blocks.

\subsection{GCA: Geometry-Constrained Attention}
\label{subsec:gca}

Skip connections in U-shaped architectures assume spatially aligned encoder--decoder tokens correspond to the same physical region, an assumption violated when anisotropic spacing makes index alignment a poor proxy for physical correspondence. GCA formulates skip interaction as geometry-constrained cross-scale alignment, decomposing affinity into a content term and a metric-derived geometric penalty on physically implausible correspondences.

For decoder token at $\mathbf{p}$ and encoder token at $\mathbf{q}$ within window $\Omega_p$:
\begin{equation}
    e_{pq}^{(\ell)} = s_{pq}^{\mathrm{cnt}} + B_{\mathrm{geo}}^{(\ell)}(\mathbf{p},\mathbf{q}),
    \label{eq:gca_logit}
\end{equation}
where $s_{pq}^{\mathrm{cnt}}$ is content similarity and $B_{\mathrm{geo}}^{(\ell)}$ is the geometric bias.

Let $\Delta_a=p_a-q_a$, $\bar{s}^{(\ell)}=(s_x^{(\ell)}s_y^{(\ell)}s_z^{(\ell)})^{1/3}$, and
$\xi_a^{(\ell)}=\frac{s_a^{(\ell)}}{\bar{s}^{(\ell)}}\Delta_a$. The geometric bias is
\begin{equation}
    B_{\mathrm{geo}}^{(\ell)}(\mathbf{p},\mathbf{q})
    = -\frac{1}{\tau^{(\ell)}}\sum_{a\in\{x,y,z\}} \big(\xi_a^{(\ell)}\big)^2
      + \sum_{a\in\{x,y,z\}} r_a^{(\ell)}\,\xi_a^{(\ell)},
    \label{eq:geo_bias}
\end{equation}
where the quadratic term acts as a metric-aware receptive field contracting attention along sparse axes and dilating along dense ones, and the linear term allows bounded directional preference. Intuitively, when slice spacing is large, $\xi_z^{(\ell)}$ grows rapidly with index offset, causing the quadratic penalty to suppress cross-slice attention to only the nearest physical neighbors. The fused output is
\begin{equation}
    \alpha_{pq}^{(\ell)}
    = \frac{\exp\!\left(e_{pq}^{(\ell)}\right)}
    {\sum_{j\in\Omega_p}\exp\!\left(e_{pj}^{(\ell)}\right)},
    \qquad
    \mathbf{y}_p
    = \sum_{q\in\Omega_p}\alpha_{pq}^{(\ell)}\,\mathbf{v}_q ,
\label{eq:gca_align}
\end{equation}
where $\mathbf{v}_q$ is the encoder value and $s_{pq}^{\mathrm{cnt}}\!=\!\langle \mathbf{q}_p,\mathbf{k}_q\rangle/\sqrt{d_k}$, yielding efficient windowed cross-attention.

We constrain parameters via
$\tau^{(\ell)}=\mathrm{softplus}(\hat{\tau}^{(\ell)})+\epsilon$ ($\epsilon=10^{-6}$) and
$r_a^{(\ell)}=r_{\max}\tanh(\hat{r}_a^{(\ell)})$,
initialized with $\tau^{(\ell)}\approx 1$ and $\hat{r}_a^{(\ell)}=0$. Parameters $\hat{\tau}^{(\ell)}$ and $\hat{r}_a^{(\ell)}$ are shared across heads within each stage.

\subsection{GVV: Geometric View Voting}
\label{subsec:gvv}

Training enforces metric consistency in representation space, but a residual geometric error persists at inference: deterministic registration of continuous anatomy onto a fixed lattice introduces systematic boundary bias beyond the reach of representation learning. GVV targets this complementary source through deterministic sub-voxel view aggregation along the anisotropic axis:
$\mathcal{D}=\{(0,0,\delta_z)\mid \delta_z\in\{0,\tfrac{1}{4},\tfrac{1}{2},\tfrac{3}{4}\}\}$,
yielding $K_z=4$ views (fixed across experiments). Each shift displaces by $\delta_z s_z^{(0)}$ mm.

Let $f_\theta(\cdot)$ denote network logits. For $\boldsymbol{\delta}\in\mathcal{D}$, $\mathcal{T}_{\boldsymbol{\delta}}$ applies trilinear sub-voxel translation and $\mathcal{T}_{-\boldsymbol{\delta}}$ maps predictions back. We aggregate:
\begin{equation}
    \bar{\mathbf{Z}}(\mathbf{x})
    =
    \frac{1}{|\mathcal{D}|}
    \sum_{\boldsymbol{\delta}\in\mathcal{D}}
    \mathcal{T}_{-\boldsymbol{\delta}}
    \left(
      f_\theta\big(\mathcal{T}_{\boldsymbol{\delta}}(\mathbf{x})\big)
    \right),
    \label{eq:gvv_logits}
\end{equation}
with $\hat{S}(\mathbf{x})=\arg\max_c \; \bar{\mathbf{Z}}_c(\mathbf{x})$.
Unlike random test-time augmentation~\cite{WANG201934}, GVV is deterministic and metric-derived: shifts and aggregation axis are prescribed by acquisition geometry.

By construction, the three components are orthogonal: MASE calibrates within-scale self-attention, GCA calibrates cross-scale skip interaction, and GVV calibrates inference-time discretization---together enforcing metric consistency wherever geometric information enters the pipeline, while remaining fully compatible with standard U-shaped Transformer backbones.

\section{Experiments and Results}
\label{sec:experiments}

\definecolor{headerblue}{RGB}{203, 204, 255} 
\definecolor{highlightblue}{RGB}{236, 234, 253} 
\definecolor{headerblue}{RGB}{203, 204, 255} 

\begin{table}[t!]
\centering
\caption{Per-organ Dice (\%,\,$\uparrow$) on BTCV and FLARE\,22 multi-organ CT segmentation.
Spl: spleen, RKi: right kidney, LKi: left kidney, Gal: gallbladder, Eso: esophagus, Liv: liver, Sto: stomach, Aor: aorta, IVC: inferior vena cava, Pan: pancreas.}
\label{tab:btcv_flare_results}
{\fontsize{8}{9.6}\selectfont
\setlength{\tabcolsep}{0.6pt}
\begin{tabular}{@{}l@{\hskip2pt}cccccccccccc@{}}
\toprule
& \textbf{Spl} & \textbf{RKi} & \textbf{LKi} & \textbf{Gal} & \textbf{Eso} & \textbf{Liv} & \textbf{Sto} & \textbf{Aor} & \textbf{IVC} & \textbf{Pan} & \textbf{AG}$^\S$ & \textbf{V/D}$^\P$ \\
\midrule
& \multicolumn{12}{c}{\cellcolor{headerblue}\textbf{BTCV}} \\
TransUNet$^\dagger$  & 87.03 & 78.60 & 81.45 & 57.96 & 66.95 & 92.97 & 75.80 & 86.42 & 71.61 & 57.62 & 47.95 & 60.92 \\
nnU-Net$^\dagger$    & 90.91 & 90.17 & 87.83 & \textbf{70.59} & 78.33 & \textbf{95.97} & 88.30 & 91.67 & 85.84 & 81.47 & 71.01 & 76.46 \\
UNETR$^\dagger$      & 83.70 & 82.41 & 79.51 & 56.59 & 64.89 & 91.70 & 75.06 & 85.01 & 75.35 & 54.43 & 53.14 & 56.86 \\
Swin-UNETR$^\dagger$ & 83.23 & 78.03 & 81.81 & 56.35 & 69.90 & 93.07 & 78.22 & 86.97 & 81.41 & 72.58 & 61.36 & 69.05 \\
U-Mamba$^\dagger$    & 90.24 & \textbf{91.20} & 89.33 & 68.63 & \textbf{79.13} & 95.25 & 87.54 & \textbf{92.85} & 86.36 & 81.95 & 71.64 & 77.05 \\
\rowcolor{highlightblue}
\textbf{Ours}        & \textbf{91.21} & 90.72 & \textbf{91.51} & 70.46 & 79.09 & 95.87 & \textbf{90.12} & 92.37 & \textbf{86.79} & \textbf{82.59} & \textbf{72.76} & \textbf{77.17} \\
\midrule
& \multicolumn{12}{c}{\cellcolor{headerblue}\textbf{FLARE\,22}} \\
TransUNet$^\dagger$  & 95.39 & 91.42 & 91.42 & 83.02 & 80.25 & 97.29 & 90.42 & 92.80 & 87.01 & 74.36 & 73.72 & 67.87 \\
nnU-Net$^\dagger$    & 98.02 & \textbf{96.84} & 97.01 & 93.45 & 89.93 & 98.93 & 95.91 & 97.28 & 95.16 & 86.91 & 89.57 & 87.54 \\
UNETR$^\dagger$      & 94.93 & 92.52 & 92.12 & 81.19 & 80.39 & 97.64 & 90.34 & 93.72 & 88.65 & 77.76 & 77.29 & 72.25 \\
Swin-UNETR$^\dagger$ & 96.01 & 94.34 & 94.61 & 88.75 & 84.71 & 97.68 & 92.23 & 94.23 & 91.13 & 82.53 & 81.14 & 81.40 \\
U-Mamba$^\dagger$    & 98.08 & 95.84 & 96.52 & \textbf{93.53} & \textbf{89.96} & \textbf{98.97} & 96.14 & \textbf{97.40} & 95.13 & 86.14 & 89.72 & 88.07 \\
\rowcolor{highlightblue}
\textbf{Ours}        & \textbf{98.87} & 96.80 & \textbf{97.65} & 93.40 & 89.58 & 98.95 & \textbf{96.73} & 97.33 & \textbf{95.17} & \textbf{87.49} & \textbf{89.76} & \textbf{89.32} \\
\bottomrule
\end{tabular}}\\
\parbox{\linewidth}{\raggedright\fontsize{8}{9.6}\selectfont
$^\dagger$ All baselines re-implemented under unified 5-fold protocol with official configurations.\\
$^\S$\,AG: mean Dice of right and left adrenal glands.\\
$^\P$\,V/D: portal and splenic veins (BTCV) or duodenum (FLARE\,22).
}
\end{table}

\subsection{Experimental Setup}
\textbf{Dataset.} We evaluate MAGEFormer on two widely used multi-organ abdominal CT benchmarks: FLARE\,22~\cite{MA2024e815} (50 labeled scans) and BTCV~\cite{btcv_2018} (30 scans). Both datasets annotate 13 abdominal organs sharing 12 common structures; dataset-specific organs are noted in Table~\ref{tab:btcv_flare_results}.\\
\noindent\textbf{Unified Evaluation Protocol.} All methods are evaluated under the same fixed 5-fold cross-validation with a shared fold split, using official model/parameter configurations and a unified metric pipeline.\\
\noindent\textbf{Implementation Details.} All models are implemented in PyTorch and trained on a single NVIDIA A100 GPU (80GB). We use AdamW with an initial learning rate of $4\times10^{-4}$, cosine annealing, and a 50-epoch warmup for 3000 epochs. Input volumes are uniformly cropped to $96\times96\times96$ with a batch size of 4.\\
\noindent\textbf{Evaluation Metrics.} We report Dice Similarity Coefficient (DSC, in \%) as the overlap metric, and the 95th percentile Hausdorff Distance (HD95, in mm) to assess boundary accuracy.

\definecolor{headerblue}{RGB}{203, 204, 255}   
\definecolor{highlightblue}{RGB}{236, 234, 253} 
\definecolor{beststd}{RGB}{89, 130, 61}          

\begin{table}[t!]
\centering
\caption{Comparison of segmentation performance on FLARE\,22 and BTCV
datasets. Results are mean$\pm$std over five
folds. Best mean values are in \textbf{bold};
\textcolor{beststd}{green} indicates the lowest standard deviation.}
\label{tab:overall_comparison}
{\fontsize{8}{9.6}\selectfont
\setlength{\tabcolsep}{2pt}
\begin{tabular}{@{} l @{\hskip 5pt} cc @{\hskip 6pt} cc @{}}
\toprule
 & \multicolumn{2}{c}{\cellcolor{headerblue}\textbf{FLARE\,22}}
 & \multicolumn{2}{c}{\cellcolor{headerblue}\textbf{BTCV}} \\
\cmidrule(lr){2-3} \cmidrule(l){4-5}
\textbf{Method}
  & \textbf{Dice}
  & \textbf{HD95}
  & \textbf{Dice}
  & \textbf{HD95} \\
\midrule
TransUNet$^\dagger$
  & \ms{84.51}{0.65}
  & \ms{15.90}{1.87}
  & \ms{70.25}{3.81}
  & \ms{28.16}{5.15} \\
nnU-Net$^\dagger$
  & \ms{93.55}{0.34}
  & \ms{4.16}{1.23}
  & \ms{83.04}{2.30}
  & \ms{13.79}{5.44} \\
UNETR$^\dagger$
  & \ms{85.85}{1.09}
  & \ms{13.39}{2.54}
  & \ms{70.14}{2.30}
  & \ms{64.13}{29.37} \\
Swin-UNETR$^\dagger$
  & \ms{89.22}{0.74}
  & \ms{31.30}{14.97}
  & \ms{74.87}{2.72}
  & \ms{75.21}{24.51} \\
U-Mamba$^\dagger$
  & \ms{93.48}{0.41}
  & \ms{4.57}{1.90}
  & \ms{83.29}{2.45}
  & \ms{11.91}{\textcolor{beststd}{\textbf{4.16}}} \\
\rowcolor{highlightblue}
\textbf{Ours}
  & \msb{93.91}{0.20}
  & \msb{3.40}{0.48}
  & \msb{84.11}{1.73}
  & \msp{10.58}{5.10} \\
\bottomrule
\end{tabular}}
\end{table}

\definecolor{highlightblue}{RGB}{236, 234, 253}
\definecolor{deltagreen}{RGB}{89, 130, 61}

\begin{table}[t]
\centering
\caption{Ablation study of proposed components (BTCV). \textbf{MASE}: Metric-Adaptive Spatial Embedding; \textbf{GCA}: Geometry-Constrained Attention; \textbf{GVV}: Geometric View Voting.}
\label{tab:ablation}
{\fontsize{8}{9.6}\selectfont
\setlength{\tabcolsep}{5pt}
\renewcommand{\arraystretch}{1.10}
\begin{tabular}{@{}lcccrr@{}}
\toprule
\textbf{Method}
  & \textbf{MASE}
  & \textbf{GCA}
  & \textbf{GVV}
  & \textbf{Avg DSC\,($\uparrow$)}
  & \textbf{HD95\,(mm,\,$\downarrow$)} \\
\midrule
Baseline
  & \textcolor{crossgray}{\ding{55}} & \textcolor{crossgray}{\ding{55}} & \textcolor{crossgray}{\ding{55}}
  & 0.7487
  & 75.21 \\
\enspace{}+ MASE
  & \checkmark & \textcolor{crossgray}{\ding{55}} & \textcolor{crossgray}{\ding{55}}
  & 0.7978\,\textcolor{deltagreen}{(+4.91\%)}
  & 37.14\,\textcolor{deltagreen}{($-$38.07)} \\
\enspace{}+ GCA
  & \textcolor{crossgray}{\ding{55}} & \checkmark & \textcolor{crossgray}{\ding{55}}
  & 0.7993\,\textcolor{deltagreen}{(+5.06\%)}
  & 34.27\,\textcolor{deltagreen}{($-$40.94)} \\
\enspace{}+ GVV
  & \textcolor{crossgray}{\ding{55}} & \textcolor{crossgray}{\ding{55}} & \checkmark
  & 0.7871\,\textcolor{deltagreen}{(+3.84\%)}
  & 62.95\,\textcolor{deltagreen}{($-$12.26)} \\
\enspace{}+ MASE + GCA
  & \checkmark & \checkmark & \textcolor{crossgray}{\ding{55}}
  & 0.8295\,\textcolor{deltagreen}{(+8.08\%)}
  & 18.76\,\textcolor{deltagreen}{($-$56.45)} \\
\rowcolor{highlightblue}
\textbf{Ours (Full)}
  & \checkmark & \checkmark & \checkmark
  & \textbf{0.8411}\,\textcolor{deltagreen}{\textbf{(+9.24\%)}}
  & \textbf{10.58}\,\textcolor{deltagreen}{\textbf{($-$64.63)}} \\
\bottomrule
\end{tabular}}
\end{table}

\begin{figure*}[t]
  \centering
  \begin{minipage}[t]{0.67\linewidth}
    \centering
    \includegraphics[width=\linewidth]{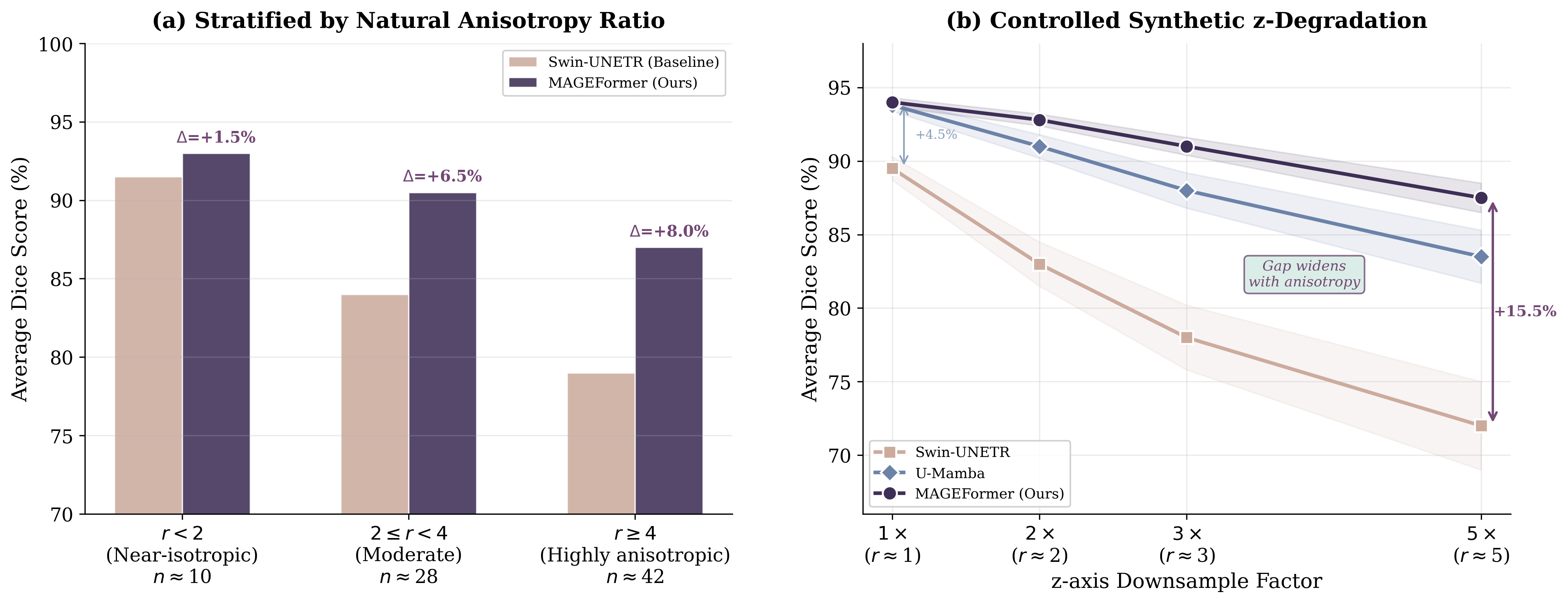}
  \end{minipage}\hfill
  \begin{minipage}[t]{0.30\linewidth}
    \centering
    \includegraphics[width=\linewidth]{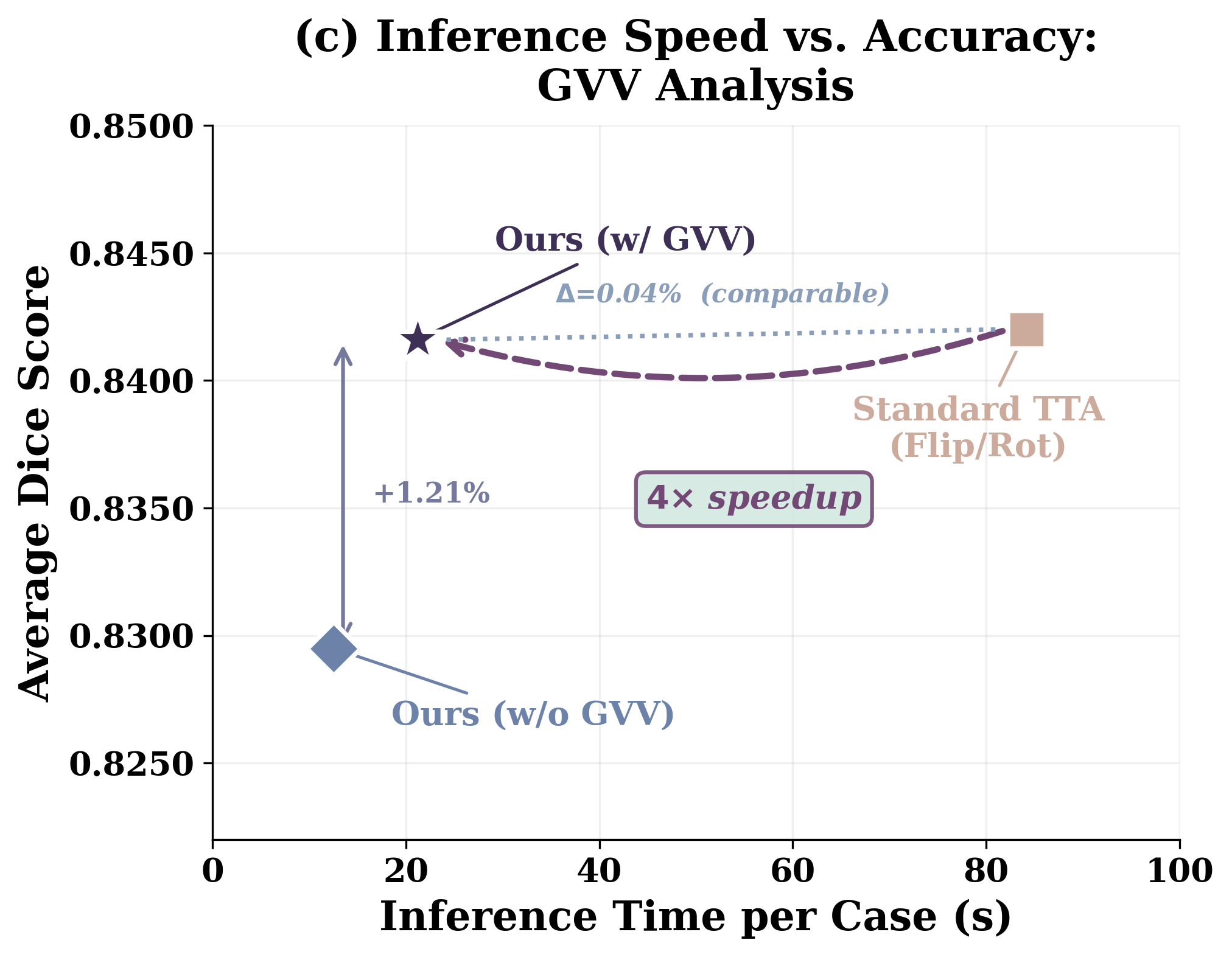}
  \end{minipage}
  \caption{Ablation and robustness analysis. (a)~Performance stratified by natural anisotropy ratio, showing increasing advantage of MAGEFormer over the baseline as anisotropy grows. (b)~Controlled synthetic z-degradation experiment demonstrating the robustness gap widens with increasing downsampling factor. (c)~Inference speed vs.\ accuracy trade-off of Geometric View Voting (GVV), achieving comparable accuracy to standard TTA with $4\times$ speedup.}
  \label{fig:ablation-analysis}
\end{figure*}
\subsection{Results}
Table~\ref{tab:overall_comparison} summarizes results against nnU-Net, TransUNet, UNETR, Swin-UNETR, and U-Mamba under the unified protocol. MAGEFormer achieves the best overall performance on both benchmarks (93.91\% Dice / 3.40\,mm HD95 on FLARE\,22; 84.11\% / 10.58\,mm on BTCV), with the strongest boundary accuracy (lowest HD95), consistent with its geometry-aware design.

Moreover, MAGEFormer exhibits the lowest cross-fold variance on both benchmarks (Dice std of 0.20 on FLARE\,22 and 1.73 on BTCV), with notably stable boundary predictions (HD95 std of 0.48 on FLARE\,22), suggesting that geometry-aware calibration yields representations that generalize more consistently across data splits.

The per-organ comparison in Table~\ref{tab:btcv_flare_results} further shows that the gains are especially evident on small or topology-sensitive organs that are more vulnerable to anisotropic sampling artifacts. On BTCV, MAGEFormer achieves the best Dice among the compared methods on the adrenal glands (AG 72.76\%) and improves pancreas segmentation to 82.59\%. These structures are highly sensitive to through-plane resolution and partial-volume effects, and the results support our claim that metric-consistent modeling improves fine-grained anatomical delineation.

The ablation on BTCV (Table~\ref{tab:ablation}, baseline: Swin-UNETR) shows that MASE and GCA each yield substantial gains individually, with marked HD95 reductions indicating improved geometric consistency. The incorporation of GVV further enhances both metrics by mitigating discretization bias. The full model achieves +9.24 Dice points (0.7487→0.8411) and reduces HD95 from 75.21 to 10.58\,mm. The robustness analysis in Fig.~\ref{fig:ablation-analysis}(a)--(b) further supports our central hypothesis: the advantage of MAGEFormer increases as anisotropy becomes more severe. In Fig.~\ref{fig:ablation-analysis}(a), pooling test cases from both benchmarks ($n=80$) and stratifying by the natural anisotropy ratio $r=s_z^{(0)}/\sqrt{s_x^{(0)}s_y^{(0)}}$ show a widening gain over the baseline as $r$ increases. In Fig.~\ref{fig:ablation-analysis}(b), under controlled synthetic z-degradation (applied at test time only by downsampling along the z-axis and interpolating back to the original grid), MAGEFormer degrades more gracefully than the compared baselines; shaded regions denote the standard deviation across folds.

\begin{figure*}[t] 
  \centering
  \includegraphics[width=\linewidth]{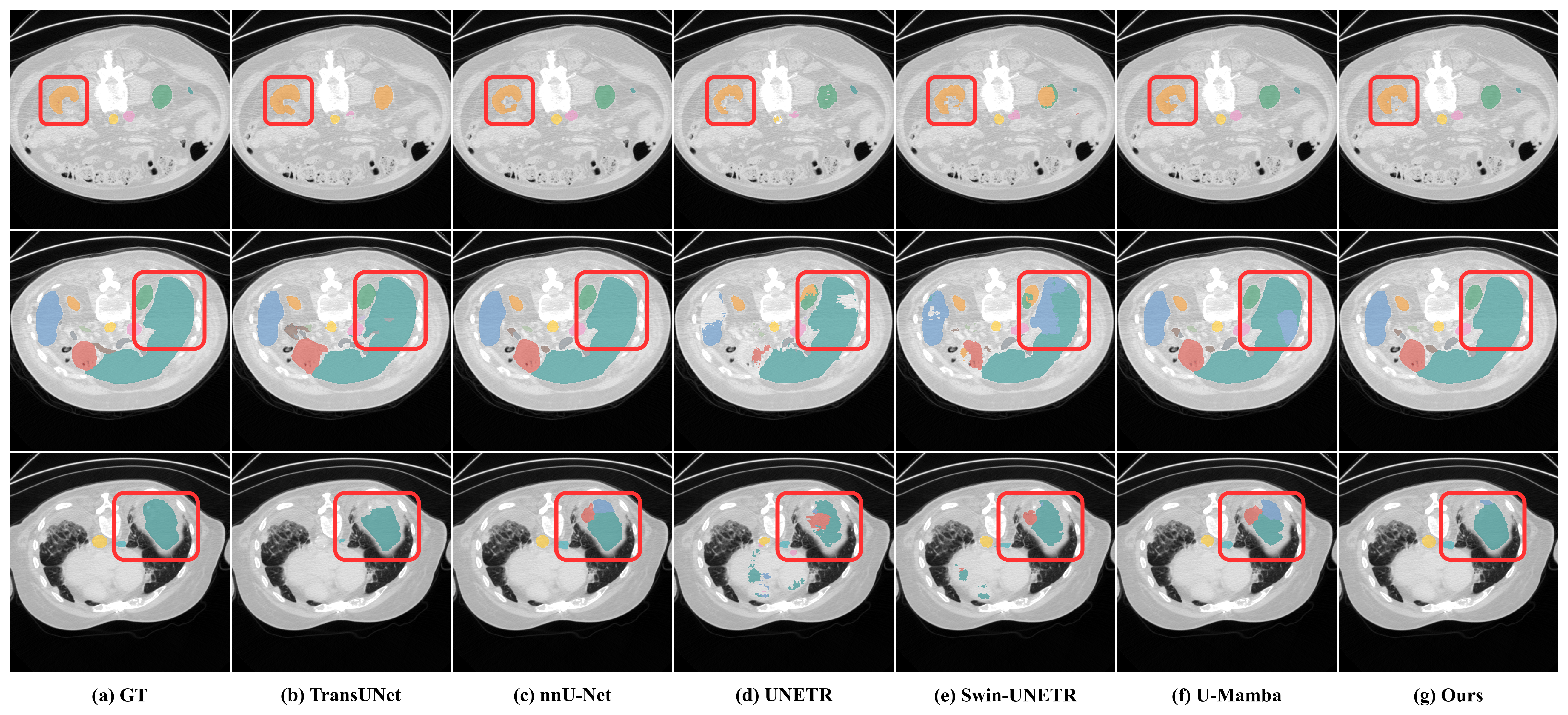}
  \caption{Qualitative visualization of different architectures. From left to right: (a) Ground Truth, (b) TransUNet~\cite{chen2024transunet}, (c) nnU-Net~\cite{nnu-net_2021}, (d) UNETR~\cite{UNETR_2022_WACV}, (e) Swin-UNETR~\cite{swinunetr22}, (f) U-Mamba~\cite{ma2024umambaenhancinglongrangedependency}, (g) Ours. Compared with baselines, MAGEFormer yields more anatomically consistent and physically plausible predictions, with less multi-label fragmentation and fewer intra-organ voids that are anatomically implausible.}
  \label{fig:view-benchmark}
\end{figure*}

Fig.~\ref{fig:ablation-analysis}(c) shows that GVV provides a favorable speed--accuracy trade-off: it achieves accuracy comparable to standard test-time augmentation (TTA) while requiring substantially lower inference time (approximately $4\times$ faster per case in our setup). Visual comparisons in Fig.~\ref{fig:view-benchmark} corroborate the quantitative findings. Compared with baseline methods, MAGEFormer produces smoother and more anatomically consistent boundaries, especially for small organs and thin structures that are prone to block-like artifacts and boundary discontinuities. The improved delineation of low-contrast regions (e.g., vascular boundaries and pancreas tail) is consistent with the role of GCA in geometry-constrained feature interaction and the effect of GVV in mitigating grid-induced discretization artifacts.

\section{Conclusion}
\label{sec:conclusion}
We presented MAGEFormer, a geometry-calibrated framework that addresses the metric mismatch between isotropic Vision Transformers and anisotropic clinical CT by embedding physical spacing constraints directly into positional encoding, cross-scale feature interaction, and inference-time view aggregation. Evaluations on the BTCV and FLARE\,22 benchmarks show that MAGEFormer achieves the strongest boundary accuracy among the compared baselines, with consistent gains in Dice, validating that internal geometry calibration is more effective than relying on conventional isotropic preprocessing alone. Future work will extend this geometry-aware paradigm to other inherently anisotropic modalities, such as multi-parametric MRI.
\begin{credits}
\subsubsection{\discintname}
The authors have no competing interests to declare that are
relevant to the content of this article.
\end{credits}

\bibliographystyle{splncs04}
\bibliography{Paper-3539}
\end{document}